\documentclass{article}

\usepackage{PRIMEarxiv}

\usepackage[utf8]{inputenc} 
\usepackage[T1]{fontenc}    
\usepackage{hyperref}       
\usepackage{url}            
\usepackage{booktabs}       
\usepackage{amsfonts}       
\usepackage{nicefrac}       
\usepackage{microtype}      
\usepackage{lipsum}
\usepackage{fancyhdr}       
\usepackage{graphicx}       
\graphicspath{{media/}}     
\usepackage{authblk}
\usepackage{graphicx,verbatim}
\usepackage{amsmath}   
\usepackage{amssymb}    
\usepackage{bm}         
\usepackage{mathtools}
\usepackage{booktabs}   
\usepackage{graphicx}
\usepackage{hyperref}
\usepackage{algorithm}
\usepackage{algorithmic}
\usepackage{multirow}
\usepackage{tabularx}
\usepackage{array}
\usepackage{cite}
\usepackage{amsmath,amssymb,amsfonts}
\usepackage{algorithmic}
\usepackage{graphicx}
\usepackage{textcomp}
\usepackage{tabularx}
\usepackage{array}
\usepackage{multirow}
\usepackage{bm}
\usepackage{adjustbox}
\usepackage{amsfonts}
\usepackage{graphicx}
\usepackage{textcomp}
\usepackage{xcolor}
\usepackage{tabularx}
\usepackage{array}
\usepackage{booktabs}
\usepackage{colortbl}
\usepackage{makecell}
\usepackage{hyperref}
\title{FreqDGT: Frequency-Adaptive Dynamic Graph Networks with Transformer for Cross-subject EEG Emotion Recognition}
\author[1]{Yueyang Li}
\author[2]{Shengyu Gong}
\author[2]{Weiming Zeng}
\author[1,*]{Nizhuan Wang}
\author[1,*]{Wai Ting Siok}

\affil[1]{Department of Chinese and Bilingual Studies, The Hong Kong Polytechnic University, Hong Kong SAR, China}
\affil[2]{School of Information Engineering, Shanghai Maritime University, Shanghai, China}

\affil[*]{Correspondence: wangnizhuan1120@gmail.com; wai-ting.siok@polyu.edu.hk}
\begin{document}
\maketitle
\begin{abstract}
Electroencephalography (EEG) serves as a reliable and objective signal for emotion recognition in affective brain-computer interfaces, offering unique advantages through its high temporal resolution and ability to capture authentic emotional states that cannot be consciously controlled. However, cross-subject generalization remains a fundamental challenge due to individual variability, cognitive traits, and emotional responses. We propose FreqDGT, a frequency-adaptive dynamic graph transformer that systematically addresses these limitations through an integrated framework. FreqDGT introduces frequency-adaptive processing (FAP) to dynamically weight emotion-relevant frequency bands based on neuroscientific evidence, employs adaptive dynamic graph learning (ADGL) to learn input-specific brain connectivity patterns, and implements multi-scale temporal disentanglement network (MTDN) that combines hierarchical temporal transformers with adversarial feature disentanglement to capture both temporal dynamics and ensure cross-subject robustness. Comprehensive experiments demonstrate that FreqDGT significantly improves cross-subject emotion recognition accuracy, confirming the effectiveness of integrating frequency-adaptive, spatial-dynamic, and temporal-hierarchical modeling while ensuring robustness to individual differences. The code is available at https://github.com/NZWANG/FreqDGT.
\end{abstract}
\section{INTRODUCTION}
Electroencephalography (EEG) has become an important tool for emotion recognition in affective brain-computer interfaces (aBCIs) \cite{kang2025hypergraph}, offering high temporal resolution to directly decode neural correlates of emotional states in real time. This non-invasive technique uniquely captures spontaneous brain electrical activities, providing objective insights into emotional processes \cite{song2018eeg}. However, cross-subject emotion recognition poses significant challenges for computational modeling due to the inherent complexity of emotional processing systems \cite{ding2024emt}. Emotions are complex psychological states that arise from integrated personal experiences, physiological responses and behavioral adaptations to environmental stimuli. At the neural level, they manifest through coordinated activity spanning multiple frequency bands — from slow $\delta$ waves involved in deep processing to fast $\gamma$ oscillations supporting cognitive integration — with each frequency range making distinct yet interactive contributions to emotional experience \cite{can2023approaches}. This complexity stems from three neurobiological challenges. The brain's functional connectivity reorganizes dynamically in response to emotional states, while individual neuroanatomical variations structurally constrain network architectures \cite{sun2022dual}. Additionally, substantial inter-individual differences in emotional cognition and expression further compound these challenges. Together, these factors pose fundamental limitations for developing robust cross-subject emotion recognition systems \cite{jin2024pgcn}.

While current graph neural networks (GNNs) have advanced spatial modeling of EEG data by treating electrodes as nodes in brain networks, their reliance on fixed adjacency matrices limits their ability to capture the dynamic connectivity patterns that characterize emotional processing \cite{gong2025lerel}. Current temporal modeling approaches assume uniform time scales, failing to account for emotions' dual temporal nature — both momentary fluctuations and sustained patterns. Most critically, existing cross-subject generalization methods compromise emotion-discriminative features when applying simplistic domain adaptation techniques to reduce inter-subject variability \cite{li2022dynamic}. These limitations necessitate an integrated framework that concurrently addresses the frequency-specific, spatially dynamic and temporally hierarchical characteristics of emotional neural processing, while maintaining individual neurobiological signatures through robust feature disentanglement. We therefore propose FreqDGT, a frequency-adaptive dynamic graph transformer that integrates adaptive frequency processing, dynamic spatial modeling, and multi-scale temporal disentanglement within a unified framework. The main contributions of this work are:
\begin{itemize}
    \item We propose a novel mechanism that learns emotion-specific frequency importance through cross-band attention and adaptive weighting, addressing the limitation that emotional states exhibit distinct frequency signatures requiring adaptive rather than uniform processing.
    
    \item We develop a multi-level adaptive dynamic graph learning approach that models emotion-dependent brain connectivity at both local and global scales. By overcoming the limitation of fixed adjacency matrices used in conventional methods, our technique accurately captures the dynamic nature of emotional brain network reorganization.
    
    \item We introduce an integrated framework that combines hierarchical temporal transformers with adversarial feature disentanglement. This architecture simultaneously models multi-scale temporal dynamics of emotional responses and explicitly separates emotion-related features from subject-specific variations. The result is significantly improved cross-subject generalization while maintaining high discriminative power for emotion classification.
\end{itemize}
\begin{figure*}[hpbt]
\centering
\includegraphics[width = 0.973\textwidth]{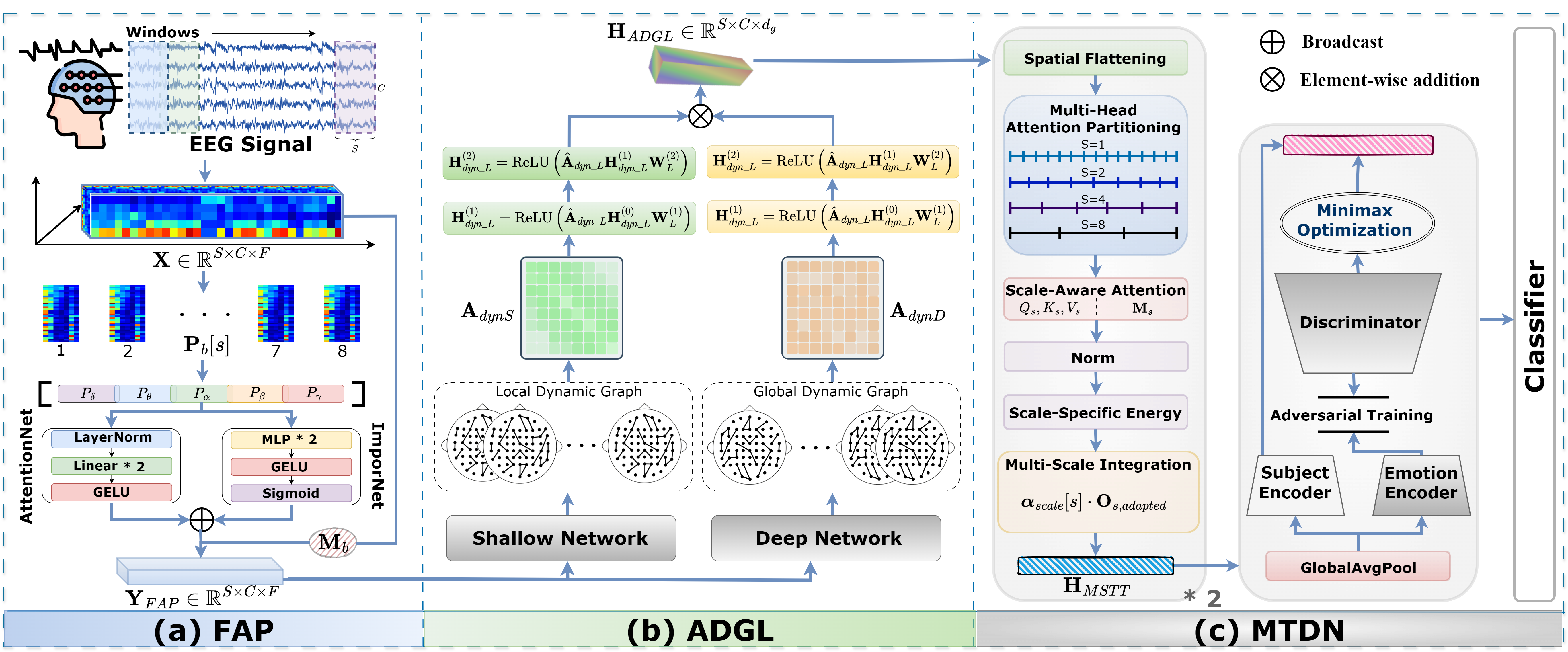}
\caption{Overall framework of the FreqDGT.}
\label{FreqDGT}
\end{figure*}
\section{RELATED WORK}
\subsection{EEG-based Emotion Recognition}
Early EEG-based emotion recognition approaches established the effectiveness of frequency-domain features such as relative power spectral density (rPSD) and differential entropy (DE) as emotional indicators \cite{ding2024emt}, though these methods typically applied uniform processing across frequency bands, failing to account for dynamic variations in spectral importance across different emotional states. The field subsequently advanced with GNNs that model the brain as interconnected networks, with DGCNN pioneering learnable adjacency matrices \cite{song2018eeg}, RGNN incorporating neurophysiological constraints \cite{zhong2020eeg}, and GCB-Net enhancing representations through broad learning systems \cite{zhang2019gcb}. More recent hybrid approaches have shown particular promise, including EEG-Conformer's combination of CNN and transformer architectures for emotion and motor imagery classification \cite{song2022eeg}, and AMDET's transformer-based attention mechanism operating across spectral-spatial-temporal dimensions \cite{xu2023amdet}. Despite these methodological innovations, current approaches continue to neglect their complex interdependencies that characterize emotional processing.

\subsection{Cross-Subject Generalization}
The inherent variability in individual neural patterns and brain structures presents significant challenges for cross-subject generalization. Early domain adaptation approaches focused on statistical alignment techniques \cite{li2022dynamic}, but these often discarded emotionally relevant information along with individual differences. Recent methodological advances have moved beyond simple alignment strategies. Meta-learning approaches such as MS-MDA develop transferable adaptation strategies to enable rapid personalization \cite{chen2021ms}, while contrastive learning approaches such as CLISA leverage the observation that responses to identical stimuli share underlying neural structures despite individual variability, employing contrastive objectives and hyperbolic embeddings respectively to identify these commonalities \cite{shen2022contrastive}. However, these approaches persist in treating inter-subject variability as noise to be minimized rather than as meaningful neurophysiological differences that could improve emotion recognition accuracy when properly modeled and incorporated.	

\section{METHODOLOGY}
We introduce FreqDGT, a frequency-adaptive dynamic graph transformer designed to overcome the challenges of cross-subject EEG emotion recognition. The framework begins with frequency-adaptive processing (FAP) that proceeds by dynamically weighting emotion-relevant bands through neuroscience-informed adaptive mechanisms. These frequency features feed into adaptive dynamic graph learning (ADGL), which addresses the fixed connectivity limitation by learning emotion-dependent spatial relationships. Subsequently, multi-scale temporal disentanglement network (MTDN) that combines hierarchical temporal processing with adversarial feature disentanglement, enabling simultaneous capture of emotion-dependent brain connectivity patterns and hierarchical temporal dynamics while ensuring cross-subject generalization.
\begin{table}
	\centering
	\caption{Key notation definitions}
	\label{tab:notation}
	\begin{tabular}{cl}
		\toprule
		\textbf{Symbol} & \textbf{Definition} \\
		\midrule
		$S, C, F$ & Number of sliding windows, channels, frequency bands \\
		$d_g, d_h, d_r$ & Graph feature, hidden, relation dimensions \\
		$\mathcal{F}_b$ & Frequency bin indices for band $b$ \\
		$\lambda, \alpha, \beta, \tau$ & Learnable/hyperparameters for attention, modulation \\
		$d_e, d_s$ & Emotion and subject feature dimensions \\
		$\mathbf{M}_b$ & Binary mask for frequency band $b$ \\
		\bottomrule
	\end{tabular}
\end{table}

\subsection{Frequency-Adaptive Processing}
Neuroscientific evidence shows distinct emotional processing roles across different frequency bands \cite{can2023approaches}. The FAP module overcomes the limitations of uniform band processing by implementing adaptive weighting mechanisms that dynamically prioritize bands according to their emotional relevance.

The module operates on rPSD features $\mathbf{X} \in \mathbb{R}^{S \times C \times F}$. For each band $b \in \{\delta, \theta, \alpha, \beta, \gamma\}$, frequency masks $\mathbf{M}_b \in \{0,1\}^F$ isolate band-specific components through element-wise multiplication. Band energy is computed through spatial aggregation to capture distributed emotional patterns:
\begin{equation}
	\mathbf{P}_b[s] = \frac{1}{C} \sum_{c=1}^{C} \sum_{f \in \mathcal{F}_b} |\mathbf{X}[s,c,f]|^2, \quad \mathbf{P}_b \in \mathbb{R}^{S}
\end{equation}

We recognize that emotional states emerge from complex interactions across frequency domains rather than isolated band activations. The concatenated band energies $\mathbf{P} = [\mathbf{P}_\delta; \mathbf{P}_\theta; \mathbf{P}_\alpha; \mathbf{P}_\beta; \mathbf{P}_\gamma] \in \mathbb{R}^{S \times 5}$ are processed through two parallel pathways, cross-band attention and importance weighting.

The cross-band attention mechanism learns inter-frequency dependencies and the importance network learns adaptive band-specific weights:
\begin{equation}
	\mathbf{A} = \text{AttentionNet}(\mathbf{P}) \in \mathbb{R}^{S \times 5}
\end{equation}
\begin{equation}
	\mathbf{W} = \text{ImporNet}(\mathbf{P}) \in \mathbb{R}^{S \times 5}
\end{equation}
where AttentionNet applies layer normalization followed by two linear transformations with GELU activation, ImporNet consists of two MLPs with GELU activation followed by sigmoid activation.

For each frequency band $b$, we apply frequency masking to obtain band-specific features:
\begin{equation}
	\mathbf{X}_b = \mathbf{M}_b \odot \mathbf{X} \in \mathbb{R}^{S \times C \times F}
\end{equation}
where $\mathbf{M}_b$ broadcasts across spatial and temporal dimensions.

The final integration combines attention and importance weights through element-wise operations:
\begin{equation}
	\mathbf{Y}_{FAP} = \sum_{i=1}^{5} \text{unsqueeze}(\mathbf{A}_{:,i} \odot \mathbf{W}_{:,i}, [1,2]) \odot \mathbf{X}_{b_i}
\end{equation}
where unsqueeze reshapes the weights to broadcast across channel and frequency dimensions, ensuring adaptive emphasis on emotion-relevant frequency patterns while maintaining signal integrity.

\subsection{Adaptive Dynamic Graph Learning} 
The fixed adjacency matrices that fail to capture the dynamic emotional brain connectivity \cite{li2022dynamic}. The ADGL module addresses this limitation through a novel multi-level dynamic graph construction approach that simultaneously learns emotion-adaptive connectivity patterns at local and global granularities.

Given the frequency features $\mathbf{Y}_{FAP} \in \mathbb{R}^{S \times C \times F}$, we first aggregate temporal information to obtain node representations suitable for graph learning:
\begin{equation}
	\mathbf{Y}'_{FAP} = \text{MeanPool}_{\text{time}}(\mathbf{Y}_{FAP}) \in \mathbb{R}^{C \times F}
\end{equation}

Emotional connectivity patterns exhibit both fine-grained local interactions and abstract global relationships, necessitating multi-granularity modeling through networks of varying representational depths. The aggregated node features are mapped through parallel relation networks with shallow and deep architectures:
\begin{equation}
	\mathbf{Z}_S = \mathcal{R}_S(\mathbf{Y}'_{FAP}) \in \mathbb{R}^{C \times d_r}
\end{equation}
\begin{equation}
	\mathbf{Z}_D = \mathcal{R}_D(\mathbf{Y}'_{FAP}) \in \mathbb{R}^{C \times d_r}
\end{equation}
where $\mathcal{R}_S(\cdot)$ captures fine-grained local patterns through shallow architecture and $\mathcal{R}_D(\cdot)$ models coarse-grained global relationships through deep transformations.

Dynamic adjacency matrices $\mathbf{A}_{dynS}$ and $\mathbf{A}_{dynD}$ are constructed through pairwise similarity computation between transformed node representations $\mathbf{Z}_S$ and $\mathbf{Z}_D$ respectively, with softmax normalization to ensure proper probability distributions. For stable propagation, matrices are symmetrized with self-connections:
\begin{equation}
	\hat{\mathbf{A}}_{dynS} = \mathbf{D}_{dynS}^{-1/2}[\frac{1}{2}(\mathbf{A}_{dynS} + \mathbf{A}_{dynS}^T) + \mathbf{I}]\mathbf{D}_{dynS}^{-1/2}
\end{equation}
\begin{equation}
	\hat{\mathbf{A}}_{dynD} = \mathbf{D}_{dynD}^{-1/2}[\frac{1}{2}(\mathbf{A}_{dynD} + \mathbf{A}_{dynD}^T) + \mathbf{I}]\mathbf{D}_{dynD}^{-1/2}
\end{equation}
where $\mathbf{D}_{dyn}[i,i] = \sum_j \tilde{\mathbf{A}}_{dyn}[i,j]$ represents the degree matrix. Multi-level graph convolution with $K$-th order Chebyshev approximation propagates features:
\begin{equation}
\mathbf{H}^{(l+1)} = \sum_{k=0}^{K-1} \theta_k^{(l)} T_k(\tilde{\mathbf{L}})\mathbf{H}^{(l)}
\end{equation}
where $T_k(\tilde{\mathbf{L}})$ denotes Chebyshev polynomials on scaled Laplacian $\tilde{\mathbf{L}} = \frac{2\mathbf{L}}{\lambda_{max}} - \mathbf{I}$ and $\theta_k^{(l)}$ are learnable parameters.

Multi-level graph convolution propagates features through normalized adjacencies:
\begin{equation}
	\mathbf{H}_{dynS}^{(l)} = \text{ReLU}(\hat{\mathbf{A}}_{dynS}\mathbf{H}_{dynS}^{(l-1)}\mathbf{W}_S^{(l)})
\end{equation}
\begin{equation}
	\mathbf{H}_{dynD}^{(l)} = \text{ReLU}(\hat{\mathbf{A}}_{dynD}\mathbf{H}_{dynD}^{(l-1)}\mathbf{W}_D^{(l)})
\end{equation}
where $\mathbf{W}_S^{(l)}, \mathbf{W}_D^{(l)} \in \mathbb{R}^{d_g \times d_g}$ are learnable transformations, initialized with $\mathbf{H}_{dynS}^{(0)} = \mathbf{H}_{dynD}^{(0)} = \mathbf{Y}'_{FAP}$.

The final representation integrates both connectivity scales and restores temporal dimension:
\begin{equation}
	\mathbf{H}_{ADGL} = \text{ExpandT}\left(\frac{1}{2}(\mathbf{H}_{dynS}^{(2)} + \mathbf{H}_{dynD}^{(2)}), S\right)
\end{equation}
where ExpandT replicates the spatial features across the temporal dimension.
\subsection{Multi-Scale Temporal Disentanglement Network}
Effective EEG emotion recognition requires simultaneous modeling of temporal dynamics across multiple scales and robust generalization across individual differences. We address these challenges through the MTDN, which integrates hierarchical temporal processing with adversarial feature disentanglement. The MTDN captures multi-scale temporal patterns through an novel transformer architecture, then separating emotion-relevant features from subject-specific variations to ensure cross-subject robustness.

\subsubsection{Multi-Scale Temporal Transformer}
Emotional responses manifest at multiple timescales, conventional transformers apply uniform attention across temporal dependencies, neglecting the hierarchical nature of emotional dynamics. Our multi-scale temporal transformer addresses this through scale-aware attention mechanisms that explicitly partition attention heads to model distinct temporal granularities within the transformer architecture.

The multi-scale temporal transformer processes spatio-temporal representations $\mathbf{H}_{ADGL} \in \mathbb{R}^{S \times C \times d_g}$ from the adaptive dynamic graph learning module while preserving spatial information through flattening, yielding $\mathbf{H}_0 = \text{Flatten}_{\text{spatial}}(\mathbf{H}_{ADGL}) \in \mathbb{R}^{S \times (C \cdot d_g)}$. Our design partitions attention heads into scale-specific groups, where each group $G_s$ operates at temporal scale $s \in \mathcal{S} = \{1, 2, 4, 8\}$.

For each scale-specific attention head group, we modify the standard attention computation to incorporate temporal scale awareness. The different head groups focus on different temporal receptive fields. Scale-specific queries, keys, and values are computed through dedicated projections, and attention weights incorporate temporal scale information:
\begin{equation}
	\mathbf{A}_{s} = \text{softmax}\left(\frac{\mathbf{Q}_{s}\mathbf{K}_{s}^T \odot \mathbf{M}_s}{\sqrt{d_k}}\right), \quad \mathbf{O}_{s} = \mathbf{A}_{s}\mathbf{V}_{s}
\end{equation}
where $\mathbf{M}_s$ represents scale-specific attention masks that constrain attention heads to focus on their designated temporal granularities, and $\odot$ denotes element-wise multiplication.

To quantify temporal pattern significance at each scale, we compute scale-specific energy measures:
\begin{equation}
	\mathbf{p}_s = \frac{1}{S}\sum_{t=1}^{S} \mathbf{O}_s[t,:] \in \mathbb{R}^{d_h}
\end{equation}

These $\mathbf{p}_s$ feed into hierarchical attention fusion that weights scale group contributions based on relevance. Scale importance weights $\boldsymbol{\alpha}_{scale}[s]$ are dynamically computed through learnable attention networks processing these measures. Adaptive modulation networks further enhance attention patterns within each scale, producing refined outputs $\mathbf{O}_{s,adapted}$.

The final multi-scale transformer output integrates all temporal scale representations:
\begin{equation}
	\mathbf{H}_{MSTT} = \text{Concat}\left(\{\boldsymbol{\alpha}_{scale}[s] \cdot \mathbf{O}_{s,adapted}\}_{s \in \mathcal{S}}\right) \mathbf{W}_O
\end{equation}
where $\mathbf{W}_O$ is a learnable projection matrix and $d_h$ represents the unified output dimension.

\subsubsection{Adversarial Disentanglement}
Cross-subject generalization faces the challenge that multi-scale temporal features $\mathbf{H}_{MSTT} \in \mathbb{R}^{S \times d_h}$ inherently contain entangled emotion-relevant information and subject-specific variations that confound model generalization. Rather than treating subject variations as noise to be eliminated, our adversarial disentanglement explicitly models these variations as structured information that can be separated from emotion-relevant features through principled feature space decomposition.

The disentanglement process begins with temporal aggregation $\mathbf{H}_{agg} = \text{GlobalAvgPool}(\mathbf{H}_{MSTT}) \in \mathbb{R}^{d_h}$, followed by dual encoding pathways with complementary objectives. The subject encoder captures individual-specific neural characteristics through learnable subject-dependent transformations following the subject-specific layer approach \cite{li2024neural}:
\begin{equation}
	\mathbf{z}_{sub} = \mathbf{M}_s \mathbf{H}_{agg} \in \mathbb{R}^{d_s}
\end{equation}
where $\mathbf{M}_s \in \mathbb{R}^{d_s \times d_h}$ represents the subject-specific transformation matrix. The emotion encoder employs a multi-layer perceptron to extract emotion-discriminative features:
\begin{equation}
	\mathbf{z}_{emo} = \mathcal{E}_{emo}(\mathbf{H}_{agg}) \in \mathbb{R}^{d_e}
\end{equation}
We make no a priori assumptions about emotion representations, instead allowing the adversarial training mechanism to guide the encoder toward learning subject-invariant emotional patterns through competitive optimization.

Adversarial training ensures orthogonal feature spaces through minimax optimization. A discriminator network $\mathcal{C}_{adv}$ attempts to predict subject identity from emotion features, while the emotion encoder is simultaneously trained to fool this discriminator. This adversarial mechanism is theoretically grounded in the principle that if emotion features cannot be used to identify subjects, they must contain minimal subject-specific information, thereby achieving the desired invariance property. The FreqDGT objective integrates all components:
\begin{equation}
	\mathcal{L}_{total} = \mathcal{L}_{cls}(\mathbf{z}_{emo}) + \lambda_{adv}\mathcal{L}_{adv} + \lambda_{disc}\mathcal{L}_{disc}
\end{equation}
where $\mathcal{L}_{cls}$ represents the emotion classification loss, $\mathcal{L}_{adv}$ denotes the adversarial loss for the emotion encoder, and $\mathcal{L}_{disc}$ represents the discriminative loss for subject identification.

	\begin{table*}[t]
	\centering\arraybackslash
	\caption{Generalized emotion classification results of different methods on the SEED, SEED-IV and FACED datasets. The best results are highlighted with dark gray background and the next best are marked with light gray background.}
	\label{tab:classification_results}
	\resizebox{\textwidth}{!}{%
		\begin{tabular}{lwc{6em}wc{6em}wc{6em}wc{6em}wc{6em}wc{6em}}
			\toprule
			\multirow{2}{*}{Method} & \multicolumn{2}{c}{SEED} & \multicolumn{2}{c}{FACED}  & \multicolumn{2}{c}{SEED-IV}              \\ \cmidrule(lr){2-3} \cmidrule(lr){4-5} \cmidrule(lr){6-7}
			& ACC (std) & F1 (std)& ACC (std) & F1 (std) & ACC (std) & F1 (std)\\ \midrule
			MLP+LSTM & 0.703 (0.172) & 0.670 (0.274) & 0.530 (0.135) &  0.619 (0.062) & 0.668 (0.063)&0.700 (0.064)\\
			DGCNN \cite{song2018eeg} & 0.710 (0.045) & 0.668 (0.251) & 
			0.551 (0.033) &  0.647 (0.052) & 0.662 (0.045) & 0.697 (0.043)\\
			GCB-Net \cite{zhang2019gcb} & 0.679 (0.162) & 0.617 (0.157) & 0.554 (0.043) &  0.620 (0.088) &0.695 (0.052) &0.685 (0.053)\\
			CLISA \cite{shen2022contrastive} & 0.771 (0.168) & 0.649 (0.106) & 0.586 (0.044) &0.677 (0.098) &0.681 (0.043)& \cellcolor{gray!20}0.731 (0.043)\\
			Conformer \cite{song2022eeg} & 0.636 (0.127) & 0.636 (0.221) & \cellcolor{gray!20}0.601 (0.038) & 0.691 (0.042)&0.690 (0.035)&0.720 (0.035)\\
			RGNN \cite{zhong2020eeg} & \cellcolor{gray!20}0.792 (0.148) & 0.802 (0.133)
			& 0.592 (0.030) &  \cellcolor{gray!20}0.738 (0.054) &0.700 (0.050)&0.722 (0.721)\\
			AMDET \cite{xu2023amdet} & 0.765 (0.140) & 0.737 (0.219) & 0.577 (0.031) &  0.720 (0.031) &0.712 (0.035)&0.673 (0.035)\\
			PGCN \cite{jin2024pgcn} &0.721 (0.168) &0.649 (0.306) & - & - &\cellcolor{gray!20}0.713 (0.043)& \cellcolor{gray!40}0.746 (0.043)\\
			EmT \cite{ding2024emt} & 0.789 (0.135) & \cellcolor{gray!40}0.821 (0.122) & 0.583 (0.049) & 0.732 (0.102) & - & - \\
			\midrule
			FreqDGT (ours) & \cellcolor{gray!40}0.811 (0.035) & \cellcolor{gray!20}0.819 (0.093) & \cellcolor{gray!40}0.623 (0.089) & \cellcolor{gray!40}0.761 (0.072) &\cellcolor{gray!40}0.719 (0.068)&0.726 (0.062)\\ \bottomrule
		\end{tabular}
	}%
\end{table*}

\section{EXPERIMENTS AND RESULTS}
	
\subsection{Datasets and Experimental Protocol}
	
We evaluate FreqDGT on three EEG emotion recognition datasets: SEED \cite{zheng2015investigating}, FACED \cite{chen2023large}, and SEED-IV \cite{zheng2018emotionmeter}.
	
\textbf{SEED} contains EEG recordings from 15 participants viewing 15 emotion-inducing film clips targeting positive, negative, and neutral states. Subsequently, the participants are tasked with providing self-evaluations regarding their emotional responses, considering dimensions such as valence and arousal. The 62-channel EEG signals were recorded at 1000 Hz and preprocessed with bandpass filtering.
	
\textbf{FACED} collected by Tsinghua University comprises 123 healthy university students including 75 females who viewed 28 emotion-eliciting video clips averaging 67 seconds across nine categories including amusement, inspiration, joy, tenderness, anger, fear, disgust, sadness and neutral emotions. The 32-channel EEG data sampled at 250 Hz from the final 30 seconds were preprocessed using bandpass filtering.
	
\textbf{SEED-IV} consists of EEG data from 15 participants (7 males and 8 females) across three sessions, with each session containing 24 trials, corresponding to 2-minute movie clips inducing four emotional states: neutral, sad, fear, and happy.
	
We employ subject-independent protocols to assess cross-subject generalization. We use leave-one-subject-out (LOSO) cross-validation with 80/20 train/validation split. SEED and FACED experiments perform binary emotion classification (positive vs. negative), with valence scores binarized at threshold 3.0 for FACED.	
\begin{figure}
	\centering
	\includegraphics[width = 0.5\textwidth]{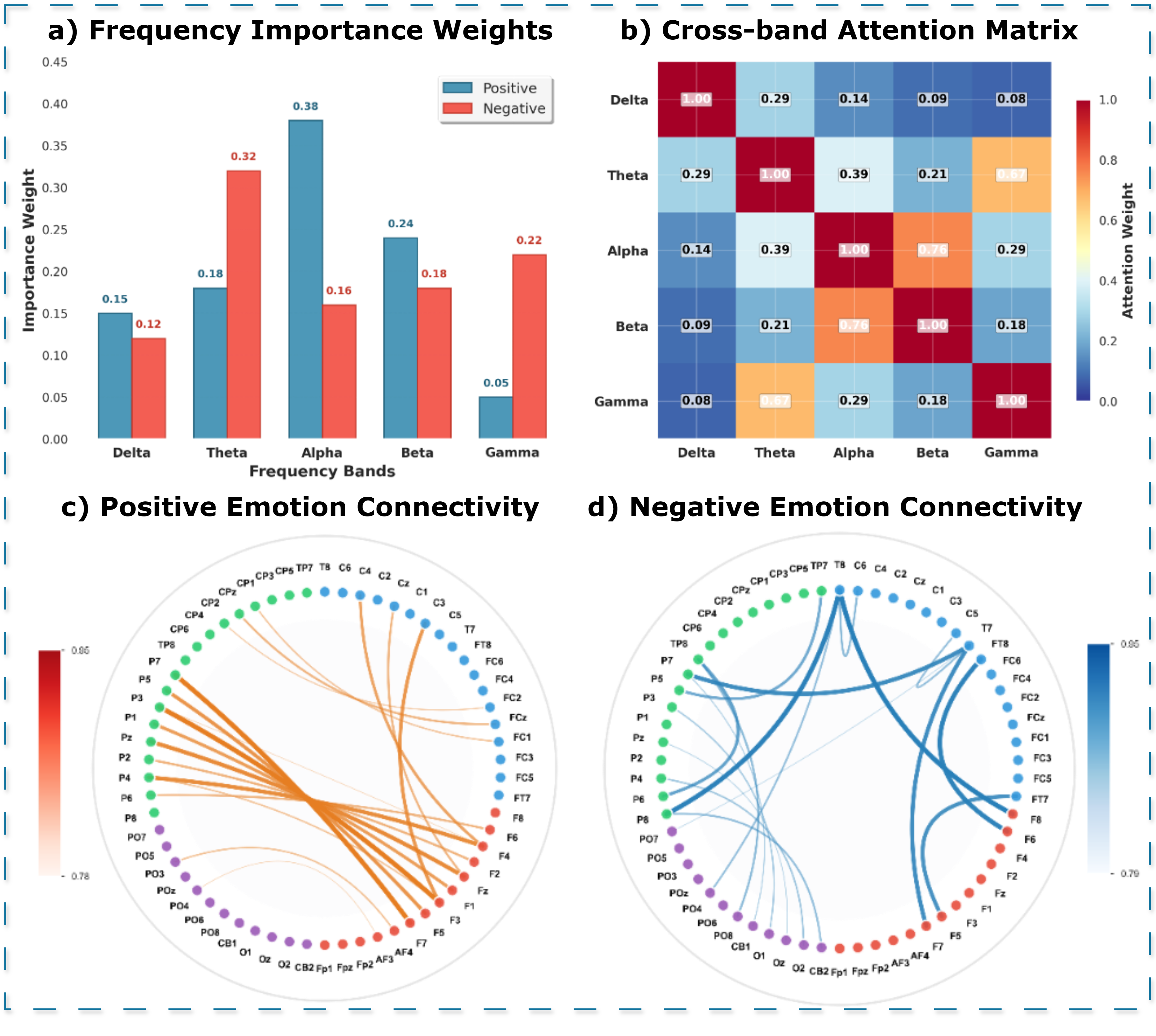}
	\caption{Visualization Analysis.}
	\label{vis}
\end{figure}

\begin{table}[!t]
	\caption{Ablation Studies with regard to Components in FreqDGT.}
	\label{table_ablation}
	\centering
	\setlength{\tabcolsep}{2.5pt}
	\renewcommand{\arraystretch}{0.9}
	\begin{tabular}{@{}c|c|ccc@{}}
		\toprule
		\multicolumn{2}{c|}{\textbf{Component Settings}} & \textbf{SEED} & \textbf{FACED} & \textbf{SEED-IV} \\
		\midrule
		\multirow{2}{*}{\makecell[c]{FAP\\Components}} & w/o Cross-band Attention & 79.3\% & 60.1\% & 69.2\% \\
		& w/o Importance Weighting & 79.8\% & 60.8\% & 69.8\% \\
		\midrule
		\multirow{2}{*}{\makecell[c]{ADGL\\Components}} & w/o Fixed Adjacency & 78.5\% & 59.4\% & 68.3\% \\
		& w/o Dynamic Learning & 79.2\% & 60.5\% & 69.1\% \\
		\midrule
		\multirow{2}{*}{\makecell[c]{MTDN\\Components}} & w/o Multi-scale Transformer & 79.1\% & 60.2\% & 69.4\% \\
		& w/o Adversarial Training & 78.6\% & 59.7\% & 68.8\% \\
        \midrule
		\multicolumn{2}{c|}{\textbf{FreqDGT (Full)}} & \textbf{81.1\%} & \textbf{62.3\%} & \textbf{71.9\%} \\
		\bottomrule
	\end{tabular}
\end{table}
\subsection{Implementation Details and Metrics}
The experiments ran on a single NVIDIA RTX 2080ti GPU. FreqDGT is implemented in PyTorch using AdamW optimizer with learning rate 5e-4 and weight decay 1e-4. The batch size is set to 128 across all datasets. The Chebyshev polynomial order $K$ is 4. To ensure reproducibility, all random seeds are fixed across experiments.
	
We report accuracy (ACC) and F1-score (F1) as standard classification metrics, measuring the proportion of correct predictions and the harmonic mean of precision and recall respectively. All results are averaged across folds with standard deviations to assess model stability.

\subsection{Overall Performance and Ablation Studies}

Table~\ref{tab:classification_results} presents comprehensive results comparing FreqDGT against state-of-the-art methods across three datasets. FreqDGT achieves the best performance with 81.1\% accuracy and 81.9\% F1-score on SEED, 62.3\% accuracy and 76.1\% F1-score on FACED, and 71.9\% accuracy and 72.6\% F1-score on SEED-IV, demonstrating substantial improvements over existing methods and validating the effectiveness of our integrated frequency-adaptive, spatial-dynamic, and temporal-hierarchical modeling approach.

Comprehensive ablation studies in Table~\ref{table_ablation} reveal the hierarchical importance of each component. MTDN emerges as the most critical module, with its removal causing the largest performance degradation across all datasets, emphasizing that multi-scale temporal modeling and feature disentanglement are fundamental for cross-subject generalization. FAP demonstrates substantial contribution, confirming that adaptive frequency processing effectively captures emotion-relevant spectral patterns. ADGL shows consistent impact, validating that dynamic graph learning captures more emotion-relevant spatial relationships than fixed connectivity approaches. Among sub-components, adversarial training within MTDN proves most significant, indicating that explicit feature disentanglement is crucial for robust cross-subject emotion recognition.

\subsection{Visualization Analysis of FreqDGT}
Figure \ref{vis} provides comprehensive insights into the neurophysiological mechanisms underlying cross-subject emotion recognition from SEED. The frequency-domain analysis reveals distinct spectral signatures between emotional states: positive emotions predominantly engage alpha and beta oscillations associated with approach behaviors and cognitive control, while negative emotions activate $\theta$ and $\gamma$ rhythms linked to threat detection and arousal regulation. The cross-frequency coupling patterns demonstrate sophisticated inter-band interactions that reflect established neurobiological processes in emotional processing \cite{ray1985eeg}.

The connectivity analysis unveils emotion-specific brain network configurations that align with contemporary neuroscientific understanding. Positive emotional states exhibit robust fronto-parietal networks with left-hemispheric preference, characteristic of approach-motivated executive control systems. Conversely, negative emotions engage temporo-frontal circuits with enhanced right-hemispheric involvement, consistent with withdrawal-related neural architectures. These findings validate FreqDGT's capacity to capture biologically plausible representations of emotional brain dynamics \cite{harmon2018role}.

\section{CONCLUSION}
We introduce FreqDGT, a frequency-adaptive dynamic graph transformer designed to overcome cross-subject EEG emotion recognition challenges through an integrated framework. By combining frequency-adaptive processing, adaptive dynamic graph learning, multi-scale temporal transformers, and adversarial disentanglement networks, FreqDGT effectively models the complex interdependencies among frequency, spatial and temporal aspects of emotional brain responses. Experimental results demonstrate significant improvements in cross-subject generalization, confirming the effectiveness of our unified approach. Future work will explore extending the framework to multimodal emotion recognition and investigating personalization strategies for practical deployment.

\section*{Acknowledgments}
This work was supported by The Hong Kong Polytechnic University Start-up Fund (Project ID: P0053210), the Hong Kong Polytechnic University Faculty Reserve Fund (Project ID: P0053738), an internal grant from The Hong Kong Polytechnic University (Project ID: P0048377), The Hong Kong Polytechnic University Departmental Collaborative Research Fund (Project ID: P0056428), The Hong Kong Polytechnic University Collaborative Research with World-leading Research Groups Fund (Project ID: P0058097) and The Research Grants Council Collaborative Research Fund (Project ID: P0049774).

\bibliographystyle{unsrt}  
\bibliography{mind2025ref}

\end{document}